\documentclass[]{oppo}
\usepackage{CJKutf8}

\usepackage{xargs}  

\usepackage{todonotes}

\usepackage{amsmath}
\usepackage{dsfont}

\usepackage{svg}
\usepackage{fontawesome}

\usepackage{array}
\usepackage{tabularx}

\usepackage{latexsym}
\usepackage{graphicx}
\usepackage[utf8]{inputenc} 
\usepackage{url}            
\usepackage{amsfonts}       
\usepackage{nicefrac}       
\usepackage{microtype}      
\usepackage{xspace}

\usepackage{listings}
\usepackage{makecell}
\usepackage{inconsolata}
\usepackage{multicol}
\usepackage{amstext}

\newcommand{\method}{ATTS}

\definecolor{darkblue}{RGB}{84, 112, 198}

\tcbset{
  aibox/.style={
    width=\linewidth,
    top=8pt,
    bottom=4pt,
    colback=blue!6!white,        
    colframe=white,              
    colbacktitle=darkblue,           
    fonttitle=\bfseries\color{white},  
    enhanced,
    center,
    attach boxed title to top left={yshift=-0.1in,xshift=0.15in},
    boxed title style={
      boxrule=0pt,               
      colframe=white,            
      colback=blue,              
    },
  }
}
\newtcolorbox{AIbox}[2][]{aibox,title=#2,#1}
\newtcolorbox{promptbox}[2][Prompt]{
colback=black!5!white,
arc=5pt, 
boxrule=0.5pt,
fonttitle=\bfseries,
title=#1, 
before upper={\small}, fontupper=\fontfamily{ptm}\selectfont,
colframe=#2 
}
\definecolor{green}{RGB}{34, 139, 34}    
\definecolor{darkblue}{RGB}{84, 112, 198}
\definecolor{Orange}{RGB}{241, 159, 77}
\renewcommand{\emph}[1]{\textit{#1}}

\DeclareRobustCommand\onedot{\futurelet\@let@token\@onedot}
\newcommand{\etc}{\emph{etc\@\onedot}}

\title{Scaling Test-time Compute for LLM Agents}
\affiliation{OPPO AI Agent Team}
\abstract{
Scaling test time compute has shown remarkable success in improving the reasoning abilities of large language models (LLMs). 
  In this work, we conduct the first systematic exploration of applying test-time scaling methods to language agents and investigate the extent to which it improves their effectiveness.
  Specifically, we explore different test-time scaling strategies, including: (1) parallel sampling algorithms; (2) sequential revision strategies; (3) verifiers and merging methods; (4)strategies for diversifying rollouts.
  We carefully analyze and ablate the impact of different design strategies on applying test-time scaling on language agents, and have follow findings:
  1. Scaling test time compute could improve the performance of agents.
2. Knowing when to reflect is important for agents.
3. Among different verification and result merging approaches, the list-wise method performs best.
4. Increasing diversified rollouts exerts a positive effect on the agent's task performance.
}

\date{\today}
\correspondence{Wangchunshu Zhou at \email{zhouwangchunshu@oppo.com}}
\checkdata[Code]{\url{https://github.com/OPPO-PersonalAI/OAgents}}

\begin{document}
\maketitle

\section{Introduction}

Language agents demonstrate exceptional capabilities in various domains~\citep{tang2025autoagent,zhou2023recurrentgpt,hong2023metagpt,smolagent2025,zhou2023agentsopensourceframeworkautonomous,zhou2024agents2}.For example, LangChain\citep{LangChain} connects LLMs with various tools to solve different tasks in an end-to-end manner, while Meta-GPT\citep{hong2023metagpt} enables multiple AI Agents to take on different roles and collaborate to accomplish tasks. Recently, long-thinking models like O1~\citep{jaech2024openai} and R1~\citep{guo2025deepseek} showcase excellent reasoning abilities of Large Language Models (LLMs). Recent approaches~\citep{owl2025,openmanus2025} leverage the extended thinking capabilities of long-activation models for planning, code writing, tool calling, and completing complex tasks. However, despite LLMs' strong capabilities, they still struggle to match human performance on complex search and reasoning tasks~\citep{zhou2023webarena,koh2024visualwebarena}. This occurs due to remaining limitations in model capabilities, errors in task planning and question answering, and issues with complex tool calling abilities. 

Increasing computational resources during the inference phase greatly enhances LLMs' performance. Some works~\citep{liu2025bagtricksinferencetimecomputation,liu20251bllmsurpass405b} improve model exploration during inference through different sampling strategies, achieving excellent scores in challenging areas like mathematical reasoning.Charlie Snell et al.\citep{snell2024scaling} investigated the effects of scaling inference-time computational consumption, while Wei Xiong et al.\citep{xiong2025self} focused on enhancing model performance through self-correction methods. However, directly applying TTS methods to the Agentic Framework presents many challenges. 
Unlike LLMs that solve specific problems in an end-to-end manner, Agents typically decompose complex problems into distinct steps, invoking multiple models sequentially for resolution. Due to the extended sequence of steps and the accumulation of errors, traditional TTS methods (e.g., BoN) can significantly undermine the final outcome, because they randomly generate N responses at each step.

To address the aforementioned challenges, we first conduct a systematic exploration of test-time scaling methods for language agents. First, we investigate the effectiveness of different \textbf{parallel sampling} methods for agentic test-time scaling, including variants of Best-of-N (BoN), beam search, and tree search algorithms. We adapt and implement these parallel sampling mechanisms within language agents and showing that despite simplicity, BoN achieves the optimal performance. 
Subsequently, we investigate the effectiveness of various \textbf{sequential revision} techniques, such as reflection and self-refinement, for language agents. We introduce a reflection agent to summarize and reflect based on the current state and recent actions/observations to help the agent consistently progress toward accomplishing the task. Experimental results show that the direct gains from having the agent perform reflection at each step are not obvious. Instead, allowing the agent to perform reflection when it performs poorly in the current step brings certain benefits. This indicates that \textbf{knowing when the agent should reflect is more important than having the agent perform reflection at every step directly}.
Finally, we conduct a detailed study on the impact of different verify and result merging methods, including voting, scoring, and list-wise approaches. Our experimental results demonstrate that whether for merge results methods or verify methods, \textbf{using the list-wise method outperforms other methods}. This provides an effective verify method reference for agentic frameworks.
Finally, we test different strategies to expand the agent's exploration space and enhance the diversity of different rollouts, and propose a multi-agent collaborative sampling strategy. Experimental results indicate that performance under multi-agent collaboration surpasses that of a single agent.

Our core contributions are:
\begin{itemize} 
\itemsep -1mm 
\item We explore the application of different parallel sampling strategies in agentic frameworks. Through parallel sampling strategies, agent performance can be significantly improved.
\item We study the impact of sequential revision techniques in agentic frameworks. In particular, we point out that it is very important for agents to know when they should perform revision.
\item We also conduct detailed comparative analysis of different verify and result merge strategies. Experiments show that the list-wise method significantly outperforms other methods.
\end{itemize} 

\section{\method: Agentic Test-Time Scaling}
In this section, we describe and compare different strategies for agentic test-time scaling including: (1) Parallel Sampling Algorithms; (2) Sequential Revision Strategies; (3) Verifier and Result Merging Methods; (4) Strategies for Diversifying Rollouts.

\subsection{Parallel Sampling Algorithms}
To establish a comprehensive evaluation framework for our proposed methods, we regard several parallel sampling algorithms that are commonly used in the test-time scaling (TTS) domain as baselines.

\textbf{Best-of-N (BoN)}
Give a sample times $N$ and question $Q$, the Best-of-N (BoN) method samples $N$ independent responses from the LLMs:
$$\{R_1, R_2, ..., R_N\} = \mbox{BoN}(Q),$$
then selects the best answer by verifying model.
The effectiveness of BoN relies heavily on the quality of the reward model and the diversity of sampled candidates.

\textbf{Step-wise-Best-of-N (BoN-wise)}
BoN selects the optimal result from candidate trajectories as the final response, while BoN-wise generates $N$ responses at each steps.
Specifically, given the thoughts $\{T1, T2, ..., T_{t-1}\}$ from the previous step at time $t$, BoN-wise generate $N$ responses: 
$$\{R_1, R_2, ..., R_N\} = \mbox{BoN-Wise}(Q, T_1, T_2, ..., T_{t-1}),$$
then it select the optimal response as thought $T_t$ at time $t$.


\textbf{Beam Search}
Beam search maintains a fixed-size beam size $K$ at each step.
Specifically, at time $t$, it generates $N$ responses for each leaf nodes $LN_i$, then maintains the most suitable $K$ responses that : 
$$\{R_1, R_2, ..., R_K\}=\mbox{BeamSearch}(Q, LN_i)$$

Most promising partial solutions at each generation step. This algorithm prunes less promising candidates early in the generation process based on cumulative log-probabilities or reward scores.

\textbf{DVTS (Diverse Verifier Tree Search)}:
DVTS decomposes the task into K subtrees, where each subtree operates as an independent beam search algorithm. By exploring multiple subtrees in parallel, DVTS achieves more diverse search behavior. Under the same computational budget, it finds higher-quality solutions more effectively than a single beam search with deep exploration.
$$\{R_1, R_2, ..., R_K\}=\mbox{DVTS}(Q, SubTree_i)$$

\begin{figure*}[t]
   \centering
   \includegraphics[width=1\textwidth]{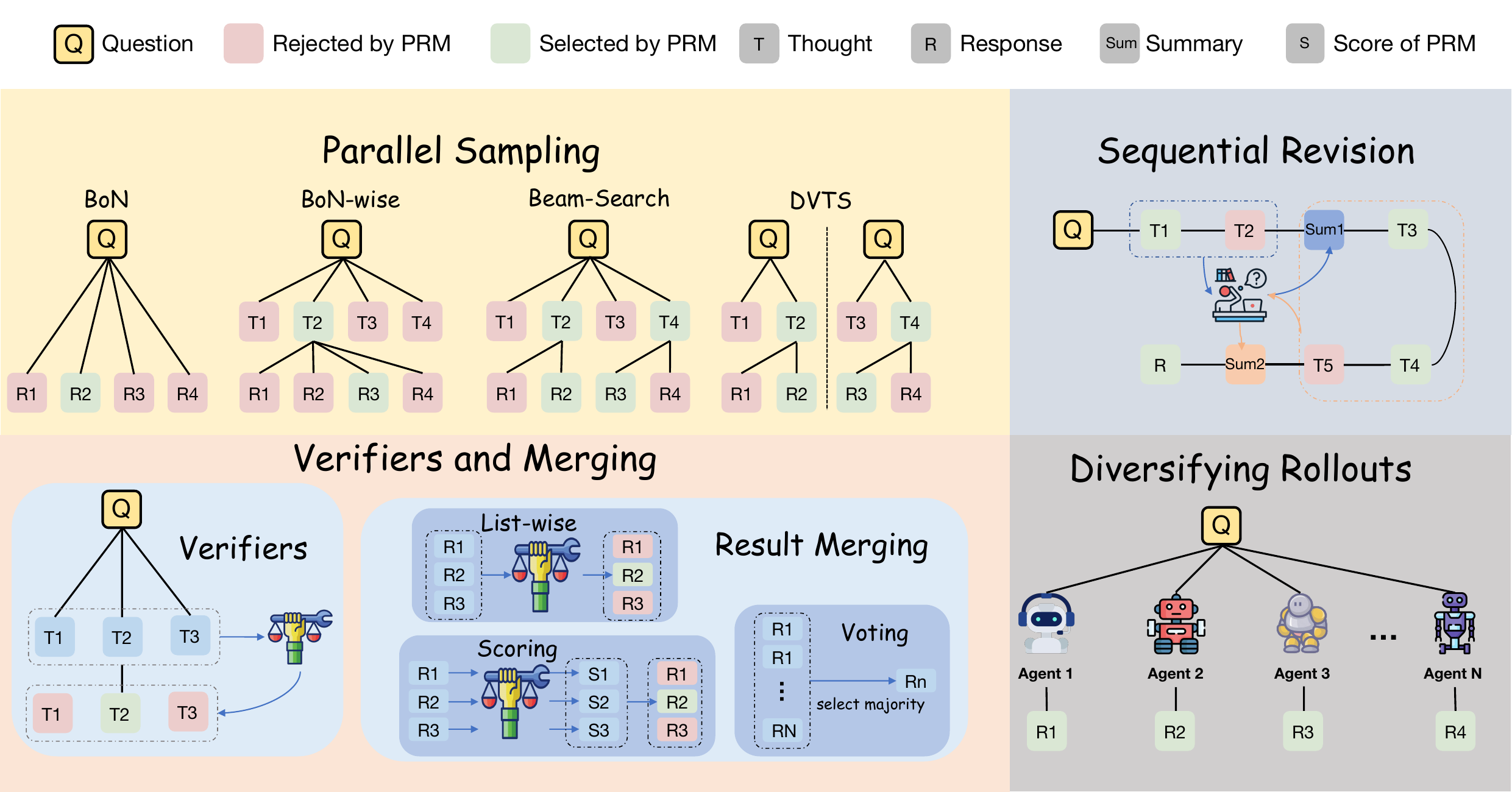}
   \caption{Overview of our agentic test-time scaling framework with four key strategies: (1) \textbf{Parallel Sampling}: BoN, BoN-wise, Beam Search, and DVTS;(2) \textbf{Sequential Revision}: Reflection model with threshold-driven re-generation;(3) \textbf{Verifiers and Result Merging Methods}: Scoring, list-wise, and majority voting;(4) \textbf{Diversifying Rollouts}: Sampling across heterogeneous agents.}  
   \label{fig:tool_exec1}          
\end{figure*}
\subsection{Sequential Revision Strategies}


Besides, given the previous steps $\{T_1, T_2, ..., T_{t-1}\}$ at time $t$, we leverage a reflection model $\mbox{RefM}$, to summarize information: $$Sum_t=\mbox{RefM}(T_1, T_2, ..., T_{t-1}).$$

To ensure the model understands when reflection is needed, we use a verify model to objectively score each step of the model to represent the quality of the current step action, and set different score  $Threshold$. If and only if the model action score is less than the $Threshold$, the $Sum_t$ is added into the LLM to generate the responses for time $t$. 



\subsection{Verifiers and Merging Methods}
\paragraph{Verifiers}To enable agents to receive positive feedback signals during the sampling process, we have designed two different process-based reward functions that evaluate the value of each sampling action. 

\textbf{scoring PRM}:
we score each thought steps at each intermediate step $t$ to revise the final response. For $N$ thought steps $\{T_{1}, T_{2}, ..., T_{N}\}$ generated at step $t$, we utilize a LLM as Reward Model ($\mbox{RM}$), to obtain the score of each response $S_i = RM(T_{current\_i})$. 

\textbf{list-wise PRM}:
Another commonly used verify method is to select the optimal trajectory through direct comparison. For $N$ thought steps $\{T_{1}, T_{2}, \ldots, T_{N}\}$ generated at step $t$, we provide all candidate actions to the LLM, asking it to select the optimal trajectory from among them, $S_i = \text{RM}(T_{\text{current\_i}})$


\paragraph{Result Merging Methods }
we compare mainstream Result Merging approaches, including \textbf{voting}: Directly select the majority from all candidates, \textbf{scoring}: using verify for direct scoring, and \textbf{list-wise}: where the model directly selects the optimal answer from candidate responses.
\subsection{Diversifying Rollouts}
The efficiency of Parallel Sampling Algorithms is influenced by diversifying rollouts—more diverse rollouts mean the agent has a greater chance of exploring and discovering the correct answer. LLMs generate diverse candidates by controlling hyperparameters such as temperature and top\_p.

However, in agent frameworks, employing multi-agents to collaboratively accomplish the same task often enhances task performance. To further increase diversity in the agents' sampling process, we utilize different LLMs as rollout models. Different LLMs often exhibit distinct capability profiles; some excel in coding, while others demonstrate exceptional performance in tool using. We have designed various agent combinations to maximize rollout diversity.

\begin{figure*}[!t] 
   \centering
\includegraphics[width=0.65\textwidth]{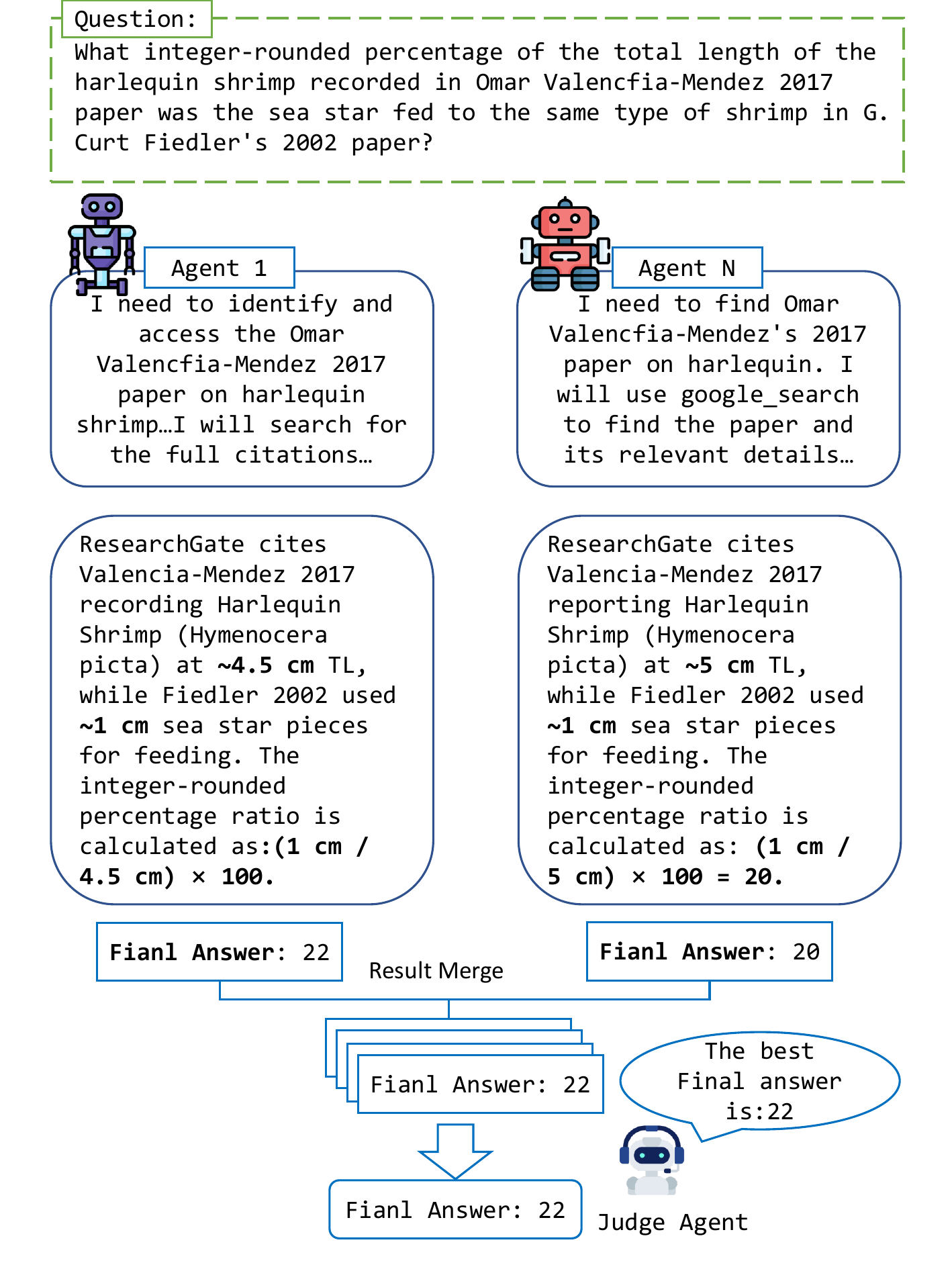}
   \caption{A case study for \method: Given one question, the agent performs operations such as coding and tool calls during a single rollout, and returns diverse results. The judge agent will merge the final result and output the best answer.}  
   \label{fig:tool_exec1}        
\end{figure*}
\section{Experiments}

\subsection{Experiments Setting}
To thoroughly investigate the effects of TTS algorithms within the agentic framework, we conduct the following comparative experiments:
\paragraph{Comparison of Different parallel sampling algorithms}
In order to comprehensively compare different parallel sampling algorithms in the agentic framework, we select mainstream parallel sampling algorithms, including BON, BON-wise, Beam-Search, Tree search, and conduct comparative experiments under identical experimental settings. We ensure a sampling width of 4, and for Beam-Search and Tree-search, we fix the beam-size at 2.
\paragraph{Comparison of different Sequential Revision Strategies}
To investigate how self-reflection affect agent performance, focusing on when and how reflection should be applied. We set up the following two settings: \textbf{Step-based Reflection}: Reflection is conducted at every step to enable continuous error correction.  \textbf{Score-based Reflection}: Initially, the Verify model scores each step of the agent's process. The agent performs reflection only when a step's score falls below a predefined threshold. To further explore how the frequency of reflection affects performance, we conduct ablation studies using three triggering thresholds: <8 (frequent), <5 (moderate), and <2 (selective).

\paragraph{Comparison of Different Verifiers and Merging Methods}
To investigate the impact of different Verifiers and Merging Methods.
First, we compare the performance differences among three mainstream result merging methods, then based on the optimal result merging method, we compare the effects of different verifiers.
\paragraph{Diversifying Rollouts}
 We explore this influence from two perspectives: on the one hand, we study the differences in agent capabilities under different sampling widths; on the other hand, we introduce multi-agent rollouts to explore the benefits of increasing rollout diversity.
\subsection{Baseline}
We select the SmoLAgents framework as our baseline. In this framework, agents take on different roles such as code actor and tool calling. In order to more intuitively compare the differences between various TTS algorithms, we remove the nesting of ToolAgent in the original smolagent framework and only use CodeAgent to directly call tools. We choose GPT-4.1 as the baseline model for the majority of our experiments. Additionally, we select current state-of-the-art models including Claude-3-7, Gemini-2.5-Pro, and Claude-3-5 for comparative experiments involving mixed models.

\subsection{Benchmark}
We choose GAIA\cite{mialon2023gaia} as our primary evaluation benchmark.
The GAIA validation dataset contains 165 data samples across three different difficulty levels - level1, level2, and level3. It primarily assesses agents' capabilities in web search and handling multimodal files. 
\section{Experimental Results}
\subsection{Comparison of Different Parallel Sampling Algorithms}
\begin{AIbox}{\makecell{Findings 1}}
\textbf{The Parallel Sampling Algorithms significantly enhance agent performance.}
\end{AIbox}
As shown in Table \ref{tab:search_Types}, we compare the application of mainstream parallel sampling algorithms in agentic frameworks. The experimental results demonstrate that by applying the parallel sampling algorithms, agents can achieve superior performance. Compared to the baseline, BoN, BoN-wise, and Beam-Search achieve significant performance gains, while DVTS performs similarly to the baseline. These results demonstrate the general effectiveness of Parallel Sampling Algorithms in the agentic framework. Meanwhile, different parallel sampling algorithms exhibit varying performance characteristics.

\begin{table*}[!ht]
  \centering
  \caption{\textbf{Comparison with Open-Source Agentic Models and Open-Source Agent Frameworks}. For the open-source models and frameworks, we adopt the results reported in their official papers. For our method, we consistently use GPT-4.1 as the base model for benchmarking.}
  \label{tab:search_Types}
    \begin{tabular}{llccccc}
    \toprule
    \midrule
    \textbf{Framework} & \textbf{Model Family} & \textbf{Average} & \textbf{Level 1} & \textbf{Level 2} & \textbf{Level 3} \\
    \midrule
    \multicolumn{5}{l}{\emph{Agentic Model}} \\
    \midrule
    Search-o1-32B~\cite{li2025search}
    & - & 39.8 & 53.8 & 34.6 & 16.7 \\
    WebThinker-32B-RL~\cite{li2025webthinker}
    & - & 48.5 & 56.4 & 50.0 & 16.7 \\
    \midrule
    \multicolumn{5}{l}{\emph{Open‐Source Agent Frameworks}} \\
    \midrule
    
    TapeAgents~\cite{bahdanau2024tapeagentsholisticframeworkagent}             
    & Claude-3-7 \etc{} & 55.76 & 71.70 & 53.49 & 30.77 \\
    AutoAgent~\cite{tang2025autoagent}              
    & Claude-3-5 \etc{} & 55.15 & 71.70 & 53.40 & 26.92 \\
    Open Deep Research~\cite{opendeepresearch}     
    & OpenAI o1 & 55.15 & 67.92 & 53.49 & 34.62 \\
    Magnetic-1~\cite{fourney2024magenticonegeneralistmultiagentsolving}             
    & OpenAI o1 \etc{} & 46.06 & 56.60 & 46.51 & 23.08 \\
    FRIDAY~\cite{wu2024copilot}                
    & GPT-4 turbo  & 34.55 & 45.28 & 34.88 & 11.54 \\
    Smolagents~\cite{smolagent2025}
    & Openai o1 \etc{} & 53.33 & 62.26 & 54.65 & 30.77 \\
    \midrule
    \multicolumn{5}{l}{\emph{Our Method}} \\
    \midrule
     Baseline   & GPT-4.1 & 55.76 & 66.04 & 58.14 & 26.92 \\
     BoN& GPT-4.1  & \textbf{63.03} & \textbf{77.36} & \textbf{63.95} & 30.77 \\
     BoN-wise & GPT-4.1  & 58.79 & 69.23 & 58.62 & \textbf{38.46} \\
     Beam-Search & GPT-4.1 & 56.97 & 69.81 & 55.81 & 34.62 \\
     DVTS & GPT-4.1 & 55.76 & 58.49 & 62.79 &26.92  \\
    \bottomrule
  \end{tabular}
\end{table*}

The BoN algorithm achieves the best performance gains, with an eight-point improvement over the baseline, and achieves SOTA results on level 1 and level 2. These two levels are heavily dependent on the agent’s ability to call and use tools. Under the BoN algorithm, the agent is given more opportunities to repeatedly attempt similar tasks, which enhances performance particularly on simpler and mid-level difficult problems. BoN-wise achieves the second-best results after BoN, with a three-point improvement over the baseline. In particular, BoN-wise achieves the best performance on the most difficult level3 problems, surpassing both the baseline and BoN. BoN-wise allows for the largest exploration space at each decision node, further demonstrating that increasing step-wise exploration leads to better performance on complex tasks.

Notably, Beam-search and DVTS show no significant improvement over baseline. This is because although these algorithms can significantly increase the agent's exploration space, their exploration also depends on the accuracy of signals provided by the verify model, which prevents the agent from stably approaching the correct answers.

\subsection{The impact of different Sequential Revision Strategies}
\begin{AIbox}{\makecell{Findings 2}}
\textbf{Understanding the opportune moments for reflection is key to its profound benefit.}
\end{AIbox}

As shown in Table \ref{tab:search_reflection}, we first compare the baseline agent with the reflection-enabled agent to assess the effectiveness of self-reflection. The baseline achieves an overall score of 55.76, while the reflection model scores slightly lower at 55.15, suggesting that reflection, while enabling error correction, may also disrupt the model’s reasoning flow. At Level 1, reflection significantly improves performance (71.7), indicating its benefit for simple tasks where minor errors can be quickly corrected without much overhead. However, at Level 2, reflection underperforms compared to the baseline, especially when applied frequently, suggesting that moderate-complexity tasks are more susceptible to disruption from excessive introspection. At Level 3, reflection leads to a moderate improvement (34.62), showing its value in preventing critical failures in complex scenarios, although overall performance remains limited by task difficulty.
\begin{table}[!ht]
  \centering
  \caption{Performance with reflection}
  \label{tab:search_reflection}
  \resizebox{0.5\textwidth}{!}{
  \begin{tabular}{lcccc}
    \toprule
      \textbf{Search type}  &\textbf{Score} & \textbf{Level 1} & \textbf{Level 2} & \textbf{Level 3} \\
    \midrule

    Baseline   & 55.76 & 66.04 & 58.14 & 26.92 \\
     Reflection   &55.15 & 71.7 & 51.16 & 34.62 \\
     \midrule
    Threshold(<8)   & 53.33 &66.04 & 53.49 & 26.92 \\
    Threshold(<5)   & 52.12 & 69.81 & 50.0 & 23.08 \\
    Threshold(<2)   & 56.36 & 71.7 & 55.81 & 26.92 \\
    \bottomrule
  \end{tabular}
  }
\end{table}

As reflection introduces both benefits and potential disruptions, we examine how varying the frequency of reflection impacts task execution across different levels of complexity. Frequent reflection (threshold <8) results in the lowest overall score (53.66), particularly hurting performance at Level 2 due to reasoning interruptions. Moderate reflection (<5) yields even lower performance (52.12), whereas selective reflection (<2) achieves the best result (56.36), outperforming other strategies across all levels. This indicates that restricting reflection to only the most critical steps minimizes disruption while still allowing meaningful error correction. These findings suggest that effective use of reflection depends heavily on its application frequency and timing—low-frequency, context-aware reflection is most beneficial, especially for maintaining coherence in multi-step reasoning processes.



\subsection{The impact of different verifiers and result merging methods}
\begin{AIbox}{\makecell{Findings 3}}
\textbf{The list-wise approach outperforms alternative methods in both verification and result merging.}
\end{AIbox}
\paragraph{The impact of different result merges method}
As shown in Table \ref{tab:search_result_merging}, we first compare common result merging methods. For these three algorithms - BoN, Beam-Search, and Tree Search - the list-wise approach outperforms other approaches. This is because: 1) compared to scoring that directly relies on standard scoring, list-wise has comparable standards for reference, making evaluation more accurate; and 2) compared to voting methods, list-wise not only considers majority options in the answers, but can also select potentially correct answers from diverse candidates.
\begin{table}[htbp]
  \centering
  \caption{Comparing performance of different result merging methods thought BoN and Beam-Search.}
  \label{tab:search_result_merging}
  \begin{tabular}{lccc}
    \toprule
      \textbf{Search type} &\textbf{voting}  & 
      \textbf{Scoring} & \textbf{list-wise}  \\
    \midrule
    BoN & 56.8 & 59.39 & 63.03 \\
    Beam-Search & 54.55 & 53.94 & 56.97 \\
    \bottomrule
  \end{tabular}
\end{table}s
\paragraph{The impact of different verify methods}
As shown in Table~\ref{tab:search_verify_method}, we compare the effects of different verify methods on agent performance. The list-wise verify method scores 3 points higher on average than the scoring, which indicate that, whether in BoN-wise or Beam-Search, using list-wise comparison of candidates is superior to the scoring approach. This suggests that compared to directly having PRM score the agent's trajectory (scoring), using a list-wise approach to have PRM select the relatively optimal trajectory can bring more precise benefits.
\begin{table}[htbp]
  \centering
  \caption{Performance with different verify methods across various search methods.}
  \label{tab:search_verify_method}
  \begin{tabular}{lcc}
    \toprule
      \textbf{Search type} & \textbf{Verify method} & 
      \textbf{Score}  \\
    \midrule
    
     BoN-wise & \makecell{scoring\\list-wise} & \makecell{56.36 \\ 58.79} \\
    \midrule
    Beam-Search & \makecell{scoring\\list-wise} & \makecell{53.94 \\ 56.97} \\
    \midrule
     Tree-Search & \makecell{scoring\\list-wise} & \makecell{50.91 \\ 55.76} \\
    \bottomrule
  \end{tabular}
\end{table}


\subsection{The impact of Diversifying Rollouts}
\begin{figure}[ht]
\centering
\includegraphics[width=0.49\textwidth]{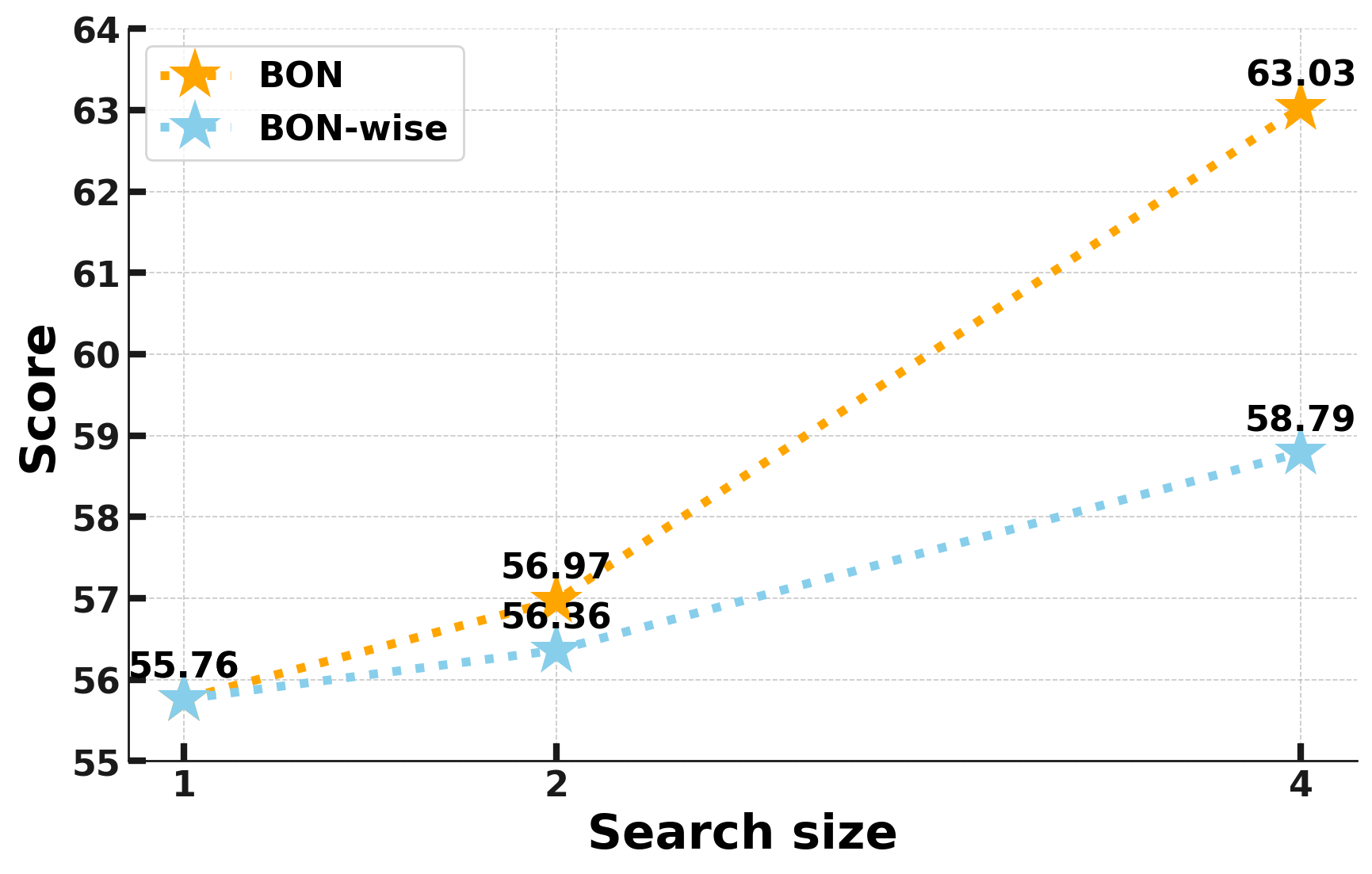} 
\caption{Performance comparison of BoN and BoN-wise algorithms across different search sizes (1, 2, and 4).}
\label{fig:performance}
\end{figure}

\begin{AIbox}{\makecell{Findings 4}}
\textbf{Increasing diversifying rollouts enhances agent performance.}
\end{AIbox}
\paragraph{Performance with different search size}
Figure~\ref{fig:performance} presents the performance variations of the agent under different sampling widths. The experimental results demonstrate that increasing the agent's sampling width leads to significant performance improvements, a finding that aligns with test-time scaling phenomena observed in the LLM domain.

\paragraph{Performance with different rollout models}
To further explore the impact of different rollout models on Diversifying Rollouts, we measure the pass@1 performance of SOTA models and calculate their pass@2 and pass@4 performance when combined with the baseline model GPT-4.1. 

As shown in Table~\ref{tab:search_performance}, under the pass@1 setting, GPT-4.1 achieves the highest score. Notably, under the pass@2 and pass@4 settings, using other models mixed with GPT-4.1 yields higher results than using GPT-4.1 alone, which further demonstrates the effectiveness of using different rollouts models. Additionally, using all four different models achieves a total score of \textbf{74.55} for pass@4, reaching a level that surpasses the open-source SOTA.

\begin{table*}[htbp]
  \centering
  \caption{Performance with different rollout models evaluated by Pass@K. Under each setting, the best performance is indicated with underlining.
  }
  \label{tab:search_performance}
\begin{tabular}{lcccc} 
    \toprule 
    Model & all & level1 & level2 & level3 \\ 
    \midrule
    GPT-4.1 & \underline{55.76} & \underline{66.04} & \underline{58.14} & 26.92 \\
    Claude-3-5 & 42.42 & 50.94 & 46.51 & 11.54 \\
    Claude-3-7 & 50.30 & 54.72 & 50.00 & \underline{42.31} \\
    Gemini-2.5-PRO & 41.82 & 54.72 & 41.86 & 15.38 \\
    \midrule
    \multicolumn{5}{l}{\emph{Pass@2}} \\
    \midrule
    GPT-4.1 only & 60.49 & 70.59 & 60.00 & 42.31 \\
    GPT-4.1,Claude-3-5 & \underline{64.24} & 71.70 & \underline{69.77} & 30.77 \\
    GPT-4.1,Claude-3-7 & \underline{64.24} & 71.70 & 63.95 & \underline{50.00} \\
    GPT-4.1,Gemini-2.5-PRO & 62.42 & \underline{79.25} & 60.47 & 34.62 \\
    \midrule
    \multicolumn{5}{l}{\emph{Pass@4}} \\
    \midrule
    GPT-4.1 only & 69.14 & 82.35 & 71.76 & 34.62 \\
    GPT-4.1,Claude-3-5, Gemini-2.5-PRO,Claude-3-7 & \underline{74.55} & \underline{86.79} & \underline{74.42} & \underline{50.00} \\
    \bottomrule 
\end{tabular}
\end{table*}
\section{Related Work}
\paragraph{Language Agents}
In previous research work, many mature agentic frameworks have been established, such as Meta-GPT\citep{hong2024metagptmetaprogrammingmultiagent} which enables GPT to assume different roles and collaboratively complete tasks, LangChain\citep{LangChain} which uses natural language to describe tools and solve complex tasks, and AGENTS\citep{zhou2023agentsopensourceframeworkautonomous} which, in addition to supporting basic tool calling and long-term memory, also supports human-agent interaction and controllability through symbolic plans (SOPs).
Recently, with the emergence of OPENAI's deep research concept, numerous agentic frameworks have appeared, such as Manus\citep{openmanus2025}, OWL\citep{owl2025}, and SmoLAgents\citep{smolagent2025}. These agentic frameworks support collaborative work among various agents, decompose task inputs, conduct multi-step task planning, and invoke diverse tools to complete complex tasks. However, current agentic frameworks predominantly employ a single linear workflow to solve problems and have not yet conducted in-depth exploration of Test-Time-Scaling (TTS) capabilities.

\paragraph{LLM Test-Time Scaling}
\citet{snell2024scaling} propose that scaling LLMs Test-time Compute\citep{wu2025inferencescalinglawsempirical} optimally can be more effective than scaling model parameters. 
OpenAI's o1 model~\footnote{\url{https://openai.com/o1/}} is designed to spend more time reasoning before they respond for the sake of obtaining better performance. 
Recently, various TTS algorithms have emerged, such as Best-Of-N, Beam-Search, Tree-Search, and Majority-Vote.\cite{liu2025can,faria2025sampledontsearchrethinking,koh2024treesearchlanguagemodel}
Moreover, \citep{kumar2024training} and \citep{xiong2025self} investigate enabling LLMs to perform self-reflection through self-rewarding and self-correlation mechanisms to enhance their performance. 

Besides, Some works(\citep{snell2024scaling, wu2024inference, huggingface2024scaling}) design reward models to guide the trajectory selection process in LLM test-time scaling . \cite{chen2025rm} incorporates deep thinking into reward models, while \cite{qian2025toolrl} uses RM for complex tool selection. 
\cite{liu20251bllmsurpass405b} and \cite{wu2024comparative} provide a comprehensive experimental analysis of LLM Test-Time Scaling. However, Test-Time Scaling strategies have not yet been fully discussed within agentic frameworks.
In this work, we investigate four key aspects of test-time scaling strategies: parallel sampling algorithms, sequential revision strategies, verifiers and merging methods, and strategies for diversifying rollouts, and conduct comprehensive ablation experiments comparing various strategies within each aspect.
\section{Conclusion}
Test Time Scaling (TTS) can significantly enhance LLM inference performance by increasing computational resources during the inference phase. However, the application of Test Time Scaling in the agentic domain still needs to be explored.
In this paper, we explore four different aspects of test-time scaling strategies: Parallel Sampling Algorithms; Sequential Revision Strategies; Verifiers and Merging Methods; Strategies for Diversifying Rollouts. We conduct detailed comparative ablation experiments for strategies in each aspect. Our experimental results indicate: 1. Applying parallel sampling algorithms to scale agent test time compute could improve agent performance, 2. For sequential revision, it is important to know when to revise, 3. Among different verify and result merge methods, list-wise methods perform best, 4. Increasing diversified rollouts exerts a positive benefit on agent, which aligns with test-time scaling phenomena observed in the LLM domain.
\newpage
\section{Contributions}

\textbf{Core Contributors}
\begin{multicols}{2}
\begin{itemize}
    \item King Zhu
    \item Hanhao Li
    \item Siwei Wu
\end{itemize}
\end{multicols}
\textbf{Contributors}
\begin{multicols}{2}
\begin{itemize}
    \item Tianshun Xing
    \item Dehua Ma
    \item Xiangru Tang
    \item Minghao Liu
    \item Jian Yang
    \item Jiaheng Liu
    \item Yuchen Eleanor Jiang
    \item Changwang Zhang
    \item Chenghua Lin
    \item Jun Wang
\end{itemize}
\end{multicols}

\textbf{Corresponding Authors}
\begin{multicols}{1}
\begin{itemize}
\item Wangchunshu Zhou
\item Ge Zhang
\end{itemize}
\end{multicols}

\clearpage
\bibliography{main}

\begin{thebibliography}{35}
\providecommand{\natexlab}[1]{#1}
\providecommand{\url}[1]{\texttt{#1}}
\expandafter\ifx\csname urlstyle\endcsname\relax
  \providecommand{\doi}[1]{doi: #1}\else
  \providecommand{\doi}{doi: \begingroup \urlstyle{rm}\Url}\fi

\bibitem[AI(2025)]{opendeepresearch}
L.~AI.
\newblock Open deep research: A fully open-source research assistant, 2025.
\newblock URL \url{https://github.com/langchain-ai/open_deep_research}.

\bibitem[Bahdanau et~al.(2024)Bahdanau, Gontier, Huang, Kamalloo, Pardinas, Piché, Scholak, Shliazhko, Tremblay, Ghanem, Parikh, Tiwari, and Vohra]{bahdanau2024tapeagentsholisticframeworkagent}
D.~Bahdanau, N.~Gontier, G.~Huang, E.~Kamalloo, R.~Pardinas, A.~Piché, T.~Scholak, O.~Shliazhko, J.~P. Tremblay, K.~Ghanem, S.~Parikh, M.~Tiwari, and Q.~Vohra.
\newblock Tapeagents: a holistic framework for agent development and optimization, 2024.
\newblock URL \url{https://arxiv.org/abs/2412.08445}.

\bibitem[Beeching et~al.(2024)Beeching, Tunstall, and Rush]{huggingface2024scaling}
E.~Beeching, L.~Tunstall, and S.~Rush.
\newblock Scaling test-time compute with open models, 2024.
\newblock URL \url{https://huggingface.co/spaces/HuggingFaceH4/blogpost-scaling-test-time-compute}.

\bibitem[Chen et~al.(2025)Chen, Li, Wang, Jin, Qian, Wang, Wang, Zhang, Zhang, Zhang, et~al.]{chen2025rm}
X.~Chen, G.~Li, Z.~Wang, B.~Jin, C.~Qian, Y.~Wang, H.~Wang, Y.~Zhang, D.~Zhang, T.~Zhang, et~al.
\newblock Rm-r1: Reward modeling as reasoning.
\newblock \emph{arXiv preprint arXiv:2505.02387}, 2025.

\bibitem[Face(2025)]{smolagent2025}
H.~Face.
\newblock Smolagent: A scalable approach to multi-agent systems, 2025.
\newblock URL \url{https://github.com/huggingface/smolagents}.

\bibitem[Faria and Smith(2025)]{faria2025sampledontsearchrethinking}
G.~Faria and N.~A. Smith.
\newblock Sample, don't search: Rethinking test-time alignment for language models, 2025.
\newblock URL \url{https://arxiv.org/abs/2504.03790}.

\bibitem[Fourney et~al.(2024)Fourney, Bansal, Mozannar, Tan, Salinas, Zhu, Niedtner, Proebsting, Bassman, Gerrits, Alber, Chang, Loynd, West, Dibia, Awadallah, Kamar, Hosn, and Amershi]{fourney2024magenticonegeneralistmultiagentsolving}
A.~Fourney, G.~Bansal, H.~Mozannar, C.~Tan, E.~Salinas, E.~E. Zhu, F.~Niedtner, G.~Proebsting, G.~Bassman, J.~Gerrits, J.~Alber, P.~Chang, R.~Loynd, R.~West, V.~Dibia, A.~Awadallah, E.~Kamar, R.~Hosn, and S.~Amershi.
\newblock Magentic-one: A generalist multi-agent system for solving complex tasks, 2024.
\newblock URL \url{https://arxiv.org/abs/2411.04468}.

\bibitem[Guo et~al.(2025)Guo, Yang, Zhang, Song, Zhang, Xu, Zhu, Ma, Wang, Bi, et~al.]{guo2025deepseek}
D.~Guo, D.~Yang, H.~Zhang, J.~Song, R.~Zhang, R.~Xu, Q.~Zhu, S.~Ma, P.~Wang, X.~Bi, et~al.
\newblock Deepseek-r1: Incentivizing reasoning capability in llms via reinforcement learning.
\newblock \emph{arXiv preprint arXiv:2501.12948}, 2025.

\bibitem[Hong et~al.(2023)Hong, Zheng, Chen, Cheng, Wang, Zhang, Wang, Yau, Lin, Zhou, et~al.]{hong2023metagpt}
S.~Hong, X.~Zheng, J.~Chen, Y.~Cheng, J.~Wang, C.~Zhang, Z.~Wang, S.~K.~S. Yau, Z.~Lin, L.~Zhou, et~al.
\newblock Metagpt: Meta programming for multi-agent collaborative framework.
\newblock \emph{arXiv preprint arXiv:2308.00352}, 3\penalty0 (4):\penalty0 6, 2023.

\bibitem[Hong et~al.(2024)Hong, Zhuge, Chen, Zheng, Cheng, Zhang, Wang, Wang, Yau, Lin, Zhou, Ran, Xiao, Wu, and Schmidhuber]{hong2024metagptmetaprogrammingmultiagent}
S.~Hong, M.~Zhuge, J.~Chen, X.~Zheng, Y.~Cheng, C.~Zhang, J.~Wang, Z.~Wang, S.~K.~S. Yau, Z.~Lin, L.~Zhou, C.~Ran, L.~Xiao, C.~Wu, and J.~Schmidhuber.
\newblock Metagpt: Meta programming for a multi-agent collaborative framework, 2024.
\newblock URL \url{https://arxiv.org/abs/2308.00352}.

\bibitem[Jaech et~al.(2024)Jaech, Kalai, Lerer, Richardson, El-Kishky, Low, Helyar, Madry, Beutel, Carney, et~al.]{jaech2024openai}
A.~Jaech, A.~Kalai, A.~Lerer, A.~Richardson, A.~El-Kishky, A.~Low, A.~Helyar, A.~Madry, A.~Beutel, A.~Carney, et~al.
\newblock Openai o1 system card.
\newblock \emph{arXiv preprint arXiv:2412.16720}, 2024.

\bibitem[Koh et~al.(2024{\natexlab{a}})Koh, Lo, Jang, Duvvur, Lim, Huang, Neubig, Zhou, Salakhutdinov, and Fried]{koh2024visualwebarena}
J.~Y. Koh, R.~Lo, L.~Jang, V.~Duvvur, M.~C. Lim, P.-Y. Huang, G.~Neubig, S.~Zhou, R.~Salakhutdinov, and D.~Fried.
\newblock Visualwebarena: Evaluating multimodal agents on realistic visual web tasks.
\newblock \emph{arXiv preprint arXiv:2401.13649}, 2024{\natexlab{a}}.

\bibitem[Koh et~al.(2024{\natexlab{b}})Koh, McAleer, Fried, and Salakhutdinov]{koh2024treesearchlanguagemodel}
J.~Y. Koh, S.~McAleer, D.~Fried, and R.~Salakhutdinov.
\newblock Tree search for language model agents, 2024{\natexlab{b}}.
\newblock URL \url{https://arxiv.org/abs/2407.01476}.

\bibitem[Kumar et~al.(2024)Kumar, Zhuang, Agarwal, Su, Co-Reyes, Singh, Baumli, Iqbal, Bishop, Roelofs, et~al.]{kumar2024training}
A.~Kumar, V.~Zhuang, R.~Agarwal, Y.~Su, J.~D. Co-Reyes, A.~Singh, K.~Baumli, S.~Iqbal, C.~Bishop, R.~Roelofs, et~al.
\newblock Training language models to self-correct via reinforcement learning.
\newblock \emph{arXiv preprint arXiv:2409.12917}, 2024.

\bibitem[Li(2025)]{owl2025}
M.~H. Y. Z. W. F. Y. N. B. X. T. S. Z. Y. Z. J. Y. L. Z. Z. Y. W. Q. Y. P. L.~G. Li.
\newblock Owl: Optimized workforce learning for general multi-agent assistance in real-world task automation, 2025.
\newblock URL \url{https://github.com/camel-ai/owl}.

\bibitem[Li et~al.(2025{\natexlab{a}})Li, Dong, Jin, Zhang, Zhou, Zhu, Zhang, and Dou]{li2025search}
X.~Li, G.~Dong, J.~Jin, Y.~Zhang, Y.~Zhou, Y.~Zhu, P.~Zhang, and Z.~Dou.
\newblock Search-o1: Agentic search-enhanced large reasoning models, 2025{\natexlab{a}}.
\newblock URL \url{https://arxiv.org/abs/2501.05366}.

\bibitem[Li et~al.(2025{\natexlab{b}})Li, Jin, Dong, Qian, Zhu, Wu, Wen, and Dou]{li2025webthinker}
X.~Li, J.~Jin, G.~Dong, H.~Qian, Y.~Zhu, Y.~Wu, J.-R. Wen, and Z.~Dou.
\newblock Webthinker: Empowering large reasoning models with deep research capability, 2025{\natexlab{b}}.
\newblock URL \url{https://arxiv.org/abs/2504.21776}.

\bibitem[Liang et~al.(2025)Liang, Xiang, Yu, Zhang, Hong, Fan, and Tang]{openmanus2025}
X.~Liang, J.~Xiang, Z.~Yu, J.~Zhang, S.~Hong, S.~Fan, and X.~Tang.
\newblock Openmanus: An open-source framework for building general ai agents, 2025.
\newblock URL \url{https://doi.org/10.5281/zenodo.15186407}.

\bibitem[Liu et~al.(2025{\natexlab{a}})Liu, Chao, Tan, and Liu]{liu2025bagtricksinferencetimecomputation}
F.~Liu, W.~Chao, N.~Tan, and H.~Liu.
\newblock Bag of tricks for inference-time computation of llm reasoning, 2025{\natexlab{a}}.
\newblock URL \url{https://arxiv.org/abs/2502.07191}.

\bibitem[Liu et~al.(2025{\natexlab{b}})Liu, Gao, Zhao, Zhang, Li, Qi, Ouyang, and Zhou]{liu20251bllmsurpass405b}
R.~Liu, J.~Gao, J.~Zhao, K.~Zhang, X.~Li, B.~Qi, W.~Ouyang, and B.~Zhou.
\newblock Can 1b llm surpass 405b llm? rethinking compute-optimal test-time scaling, 2025{\natexlab{b}}.
\newblock URL \url{https://arxiv.org/abs/2502.06703}.

\bibitem[Liu et~al.(2025{\natexlab{c}})Liu, Gao, Zhao, Zhang, Li, Qi, Ouyang, and Zhou]{liu2025can}
R.~Liu, J.~Gao, J.~Zhao, K.~Zhang, X.~Li, B.~Qi, W.~Ouyang, and B.~Zhou.
\newblock Can 1b llm surpass 405b llm? rethinking compute-optimal test-time scaling.
\newblock \emph{arXiv preprint arXiv:2502.06703}, 2025{\natexlab{c}}.

\bibitem[Mialon et~al.(2023)Mialon, Fourrier, Wolf, LeCun, and Scialom]{mialon2023gaia}
G.~Mialon, C.~Fourrier, T.~Wolf, Y.~LeCun, and T.~Scialom.
\newblock Gaia: a benchmark for general ai assistants.
\newblock In \emph{The Twelfth International Conference on Learning Representations}, 2023.

\bibitem[Qian et~al.(2025)Qian, Acikgoz, He, Wang, Chen, Hakkani-T{\"u}r, Tur, and Ji]{qian2025toolrl}
C.~Qian, E.~C. Acikgoz, Q.~He, H.~Wang, X.~Chen, D.~Hakkani-T{\"u}r, G.~Tur, and H.~Ji.
\newblock Toolrl: Reward is all tool learning needs.
\newblock \emph{arXiv preprint arXiv:2504.13958}, 2025.

\bibitem[repository(2022)]{LangChain}
L.~repository.
\newblock Langchain, 2022.
\newblock URL \url{https://github.com/langchain-ai/langchain.}

\bibitem[Snell et~al.(2024)Snell, Lee, Xu, and Kumar]{snell2024scaling}
C.~Snell, J.~Lee, K.~Xu, and A.~Kumar.
\newblock Scaling llm test-time compute optimally can be more effective than scaling model parameters.
\newblock \emph{arXiv preprint arXiv:2408.03314}, 2024.

\bibitem[Tang et~al.(2025)Tang, Fan, and Huang]{tang2025autoagent}
J.~Tang, T.~Fan, and C.~Huang.
\newblock Autoagent: A fully-automated and zero-code framework for llm agents, 2025.
\newblock URL \url{https://arxiv.org/abs/2502.05957}.

\bibitem[Wu et~al.(2024{\natexlab{a}})Wu, Peng, Du, Zheng, Liu, Wu, Ma, Li, Yang, Zhou, et~al.]{wu2024comparative}
S.~Wu, Z.~Peng, X.~Du, T.~Zheng, M.~Liu, J.~Wu, J.~Ma, Y.~Li, J.~Yang, W.~Zhou, et~al.
\newblock A comparative study on reasoning patterns of openai's o1 model.
\newblock \emph{arXiv preprint arXiv:2410.13639}, 2024{\natexlab{a}}.

\bibitem[Wu et~al.(2024{\natexlab{b}})Wu, Sun, Li, Welleck, and Yang]{wu2024inference}
Y.~Wu, Z.~Sun, S.~Li, S.~Welleck, and Y.~Yang.
\newblock Inference scaling laws: An empirical analysis of compute-optimal inference for problem-solving with language models.
\newblock \emph{arXiv preprint arXiv:2408.00724}, 2024{\natexlab{b}}.

\bibitem[Wu et~al.(2025)Wu, Sun, Li, Welleck, and Yang]{wu2025inferencescalinglawsempirical}
Y.~Wu, Z.~Sun, S.~Li, S.~Welleck, and Y.~Yang.
\newblock Inference scaling laws: An empirical analysis of compute-optimal inference for problem-solving with language models, 2025.
\newblock URL \url{https://arxiv.org/abs/2408.00724}.

\bibitem[Wu et~al.(2024{\natexlab{c}})Wu, Han, Ding, Weng, Liu, Yao, Yu, and Kong]{wu2024copilot}
Z.~Wu, C.~Han, Z.~Ding, Z.~Weng, Z.~Liu, S.~Yao, T.~Yu, and L.~Kong.
\newblock Os-copilot: Towards generalist computer agents with self-improvement, 2024{\natexlab{c}}.
\newblock URL \url{https://arxiv.org/abs/2402.07456}.

\bibitem[Xiong et~al.(2025)Xiong, Zhang, Ye, Chen, Jiang, and Zhang]{xiong2025self}
W.~Xiong, H.~Zhang, C.~Ye, L.~Chen, N.~Jiang, and T.~Zhang.
\newblock Self-rewarding correction for mathematical reasoning.
\newblock \emph{arXiv preprint arXiv:2502.19613}, 2025.

\bibitem[Zhou et~al.(2023{\natexlab{a}})Zhou, Xu, Zhu, Zhou, Lo, Sridhar, Cheng, Ou, Bisk, Fried, et~al.]{zhou2023webarena}
S.~Zhou, F.~F. Xu, H.~Zhu, X.~Zhou, R.~Lo, A.~Sridhar, X.~Cheng, T.~Ou, Y.~Bisk, D.~Fried, et~al.
\newblock Webarena: A realistic web environment for building autonomous agents.
\newblock \emph{arXiv preprint arXiv:2307.13854}, 2023{\natexlab{a}}.

\bibitem[Zhou et~al.(2023{\natexlab{b}})Zhou, Jiang, Cui, Wang, Xiao, Hou, Cotterell, and Sachan]{zhou2023recurrentgpt}
W.~Zhou, Y.~E. Jiang, P.~Cui, T.~Wang, Z.~Xiao, Y.~Hou, R.~Cotterell, and M.~Sachan.
\newblock Recurrentgpt: Interactive generation of (arbitrarily) long text, 2023{\natexlab{b}}.
\newblock URL \url{https://arxiv.org/abs/2305.13304}.

\bibitem[Zhou et~al.(2023{\natexlab{c}})Zhou, Jiang, Li, Wu, Wang, Qiu, Zhang, Chen, Wu, Wang, Zhu, Chen, Zhang, Tang, Zhang, Chen, Cui, and Sachan]{zhou2023agentsopensourceframeworkautonomous}
W.~Zhou, Y.~E. Jiang, L.~Li, J.~Wu, T.~Wang, S.~Qiu, J.~Zhang, J.~Chen, R.~Wu, S.~Wang, S.~Zhu, J.~Chen, W.~Zhang, X.~Tang, N.~Zhang, H.~Chen, P.~Cui, and M.~Sachan.
\newblock Agents: An open-source framework for autonomous language agents, 2023{\natexlab{c}}.
\newblock URL \url{https://arxiv.org/abs/2309.07870}.

\bibitem[Zhou et~al.(2024)Zhou, Ou, Ding, Li, Wu, Wang, Chen, Wang, Xu, Zhang, Chen, and Jiang]{zhou2024agents2}
W.~Zhou, Y.~Ou, S.~Ding, L.~Li, J.~Wu, T.~Wang, J.~Chen, S.~Wang, X.~Xu, N.~Zhang, H.~Chen, and Y.~E. Jiang.
\newblock Symbolic learning enables self-evolving agents.
\newblock 2024.
\newblock URL \url{https://arxiv.org/abs/2406.18532}.

\end{thebibliography}
\section{Appendix}
\scriptsize 
\begin{promptbox}[PRM-score Evaluation Prompt]{green}
\textbf{Evaluation Guidelines:}
\begin{itemize}
    \item \textbf{Objective:}
    \begin{itemize}
        \item You will evaluate a candidate ActionStep node, which includes the following fields:
        \begin{itemize}
            \item \texttt{step\_number}: Depth of this step within the TTS search tree.
            \item \texttt{observations}: Observations recorded after executing this action.
            \item \texttt{action\_output}: Direct output resulting from this action.
            \item \texttt{model\_output}: Raw LLM output that led to this action.
            \item \texttt{error}: Any encountered errors (can be None).
            \item \texttt{score}: Previously assigned score (for reference only).
            \item \texttt{previous\_steps}: The history of earlier steps, including TaskStep and PlanningStep, along with the trajectory of ActionSteps leading to the current state.
        \end{itemize}
        \item Your goal is to judge how promising this ActionStep is for advancing toward the user's task, using your independent judgment while considering the continuity and logical flow of the ActionStep sequence, including the historical context.
    \end{itemize}

    \item \textbf{Evaluation Criteria:}
    \begin{itemize}
        \item \textbf{Progress Toward Goal:}
        \begin{itemize}
            \item Assess whether the \texttt{action\_output} clearly and tangibly advances the overall task.
            \item Reward meaningful progress or valuable new information.
            \item Penalize irrelevant actions or weak impact.
        \end{itemize}
        \item \textbf{Error and Stability:}
        \begin{itemize}
            \item Penalize based on the severity of errors:
            \begin{itemize}
                \item Fatal/blocking errors: 0-1 points.
                \item Significant errors: 1-3 points.
                \item Minor or recoverable errors: 3-5 points.
            \end{itemize}
            \item Reduce the score if the \texttt{model\_output} is ambiguous or unstable.
        \end{itemize}
        \item \textbf{TTS Efficiency:}
        \begin{itemize}
            \item Reward actions that contribute efficiently toward reaching the goal.
            \item Penalize redundant or repetitive actions without meaningful progress.
        \end{itemize}
        \item \textbf{Reflection Usage:}
        \begin{itemize}
            \item Reward active utilization of \texttt{reflection} to improve upon past mistakes.
            \item Penalize ignoring reflection insights.
        \end{itemize}
        \item \textbf{Loop Detection:}
        \begin{itemize}
            \item Detect loops or repetitions compared to previous steps.
            \item Identify true loops and penalize based on severity.
        \end{itemize}
        \item \textbf{Contextual Awareness:}
        \begin{itemize}
            \item Infer alignment with previous \texttt{PlanningStep} and \texttt{TaskStep}.
            \item Ensure consistency with the TTS strategy and penalize deviations.
        \end{itemize}
    \end{itemize}

    \item \textbf{Scoring Criteria:}
    \begin{itemize}
        \item \texttt{9-10}: Clearly advances the goal; highly efficient; strong reflection use; no loops.
        \item \texttt{7-8}: Good advancement; minor inefficiencies; clear reflection use; minimal loop risk.
        \item \texttt{5-6}: Moderate progress; limited efficiency; moderate reflection use; mild repetition risks.
        \item \texttt{3-4}: Poor advancement; inefficient; weak reflection use; noticeable loop risks.
        \item \texttt{1-2}: Minimal advancement; repetitive actions; true loops; significant errors.
        \item \texttt{0}: Severe issues: explicit loops, critical errors, or complete irrelevance to the task context.
    \end{itemize}

    \item \textbf{Final Evaluation Output:}
    You must provide your evaluation in valid JSON format with the following structure:
    \begin{quote}
    \texttt{
    \{
  "analysis": "Detailed analysis addressing     progress, TTS efficiency, reflection usage, loop detection, contextual alignment with PlanningStep/TaskStep, error severity, and overall action quality.",
  "score": [integer between 0-10]
    \}
    }
    \end{quote}
    \end{itemize}
\end{promptbox}

\begin{promptbox}[PRM-list Evaluation Prompt]{green}
\scriptsize 
\textbf{Evaluation Guidelines:}
\begin{itemize}
    \item \textbf{Objective:}
    \begin{itemize}
        \item You will evaluate N candidate trajectories, each representing a series of nodes in a search tree. Each trajectory contains the following:
        \begin{itemize}
            \item \texttt{step\_number}: Depth of the node in the trajectory.
            \item \texttt{observations}: Observations recorded at each step of the trajectory.
            \item \texttt{action\_output}: Direct action output at each step.
            \item \texttt{model\_output}: Raw model output (LLM).
            \item \texttt{error}: Any errors encountered (can be None).
            \item \texttt{score}: Previously calculated score (if available).
            \item \texttt{previous\_steps}: The history of earlier steps, including TaskStep and PlanningStep, with the trajectory of ActionSteps leading to the current state.
        \end{itemize}
        \item Your goal is to evaluate each trajectory holistically, considering how well it progresses toward solving the user's task. Select the trajectory that most effectively achieves this goal.
    \end{itemize}

    \item \textbf{Evaluation Criteria:}
    \begin{itemize}
        \item \textbf{Progress Toward Goal:}
        \begin{itemize}
            \item Assess how well each trajectory advances the task at hand, considering both the individual node's progress and the overall progression of the entire trajectory.
            \item Reward trajectories that demonstrate tangible and meaningful progress toward the goal.
            \item Penalize trajectories with weak actions or minimal/no advancement.
        \end{itemize}
        \item \textbf{Trajectory Efficiency:}
        \begin{itemize}
            \item Evaluate how efficiently each trajectory progresses toward the goal, considering the depth and complexity of the steps.
            \item Favor trajectories that achieve significant progress with fewer steps.
            \item Consider the overall value-to-depth ratio when comparing trajectories of different lengths.
            \item Reward efficient exploration of the search space.
        \end{itemize}
        \item \textbf{Loop Detection:}
        \begin{itemize}
            \item Detect loops or repetitions within each trajectory, especially those related to previous steps.
            \item \textbf{Loop types:}
            \begin{itemize}
                \item \texttt{Real Loops}: Identical nodes (observations, action output, and model output) that do not add value to the trajectory.
                \item \texttt{Benign Repetitions}: Similar strategies with variations yielding additional progress.
            \end{itemize}
            \item Heavily penalize trajectories with real loops.
            \item Slight penalties for benign repetitions if they lead to meaningful improvements.
        \end{itemize}
        \item \textbf{Error and Stability:}
        \begin{itemize}
            \item Evaluate the severity of errors encountered in each trajectory and penalize based on their impact on progression.
            \item \textbf{Error Severity:}
            \begin{itemize}
                \item Fatal/Blocking Errors: Major penalty.
                \item Significant Errors: Moderate penalty.
                \item Minor/Recoverable Issues: Minor penalty.
            \end{itemize}
            \item Penalize unstable or unclear model outputs.
            \item Consider how errors affect the overall trajectory's ability to move toward the goal.
        \end{itemize}
        \item \textbf{Overall Trajectory Quality:}
        \begin{itemize}
            \item Evaluate the coherence and overall quality of the trajectory.
            \item Consider the logical sequence of steps and the exploration-exploitation balance.
            \item Evaluate the final node's closeness to achieving the goal.
            \item Reward trajectories that make consistent progress and demonstrate coherent planning.
        \end{itemize}
    \end{itemize}

    \item \textbf{Final Output Format:}
    Provide your evaluation in the following JSON format. Select the best trajectory and provide a detailed analysis explaining why it is the most promising trajectory.
    \begin{quote}
    \texttt{
    \{
  "index": [integer],  \# Index of the best trajectory
  "analysis": "Detailed analysis addressing progress, efficiency, reflection usage, loop detection, error severity, and overall trajectory quality."
    \}
    }
    \end{quote}
    \end{itemize}

\end{promptbox}

\label{sec:appendix}

\begin{promptbox}[Single Node Reflection Prompt]{blue}

\textbf{Node Information:}
\begin{itemize}
  \item \texttt{step\_number}: The depth of the node within the BON/beam search tree.
  \item \texttt{observations}: The data or observations recorded during this step.
  \item \texttt{action\_output}: The direct output resulting from an action taken at this step (e.g., API call, tool response).
  \item \texttt{model\_output}: The raw output generated by the model at this step.
  \item \texttt{error}: Any errors encountered during this step (if applicable).
\end{itemize}

\textbf{Goal:}
\begin{itemize}
  \item \textbf{Summarize:}
  \begin{itemize}
    \item Provide a brief overview of what occurred at this node.
    \item Describe the action taken and the results or new information that emerged as a result of this action.
  \end{itemize}

  \item \textbf{Reflect:}
  \begin{itemize}
    \item Assess whether the action taken in this node was successful, partially successful, or unsuccessful.
    \item Identify any errors, issues, or incompleteness relevant to this step.
    \item Compare the node’s outcome with its assigned score, providing an evaluation of whether the score is aligned with the actual result.
  \end{itemize}

  \item \textbf{Confidence:}
  \begin{itemize}
    \item Evaluate your confidence in the action taken at this node (High/Medium/Low).
    \item If confidence is high, explicitly suggest continuing along this exploration path.
    \item If confidence is medium or low, recommend potential improvements or alternatives, while leaving room for exploration to remain open.
  \end{itemize}

  \item \textbf{Suggest:}
  \begin{itemize}
    \item Provide specific and focused suggestions for refining the current step.
    \item These should be based on the evaluation of the current node, with an emphasis on actionable changes that can be made in the next attempt of a similar step.
    \item Focus exclusively on improvements that can be applied within this node. Avoid proposing changes that span multiple steps or introduce larger, long-term strategies.
    \item Base your evaluation strictly on the provided fields—action\_output, observations, error, etc. Do not infer additional context or hypothesize about alternative paths or unknown factors.
    \item Only flag a step as unsuccessful or in need of improvement if there is clear, tangible evidence (e.g., explicit errors, missing or incorrect outputs).
    \item Do not override factual results based on subjective judgment, even if the node's score does not seem to match the outcome.
  \end{itemize}

  \item \textbf{General Guidelines:}
  \begin{itemize}
    \item Your suggestions should be conservative, focusing only on changes where there is a clear issue or opportunity for improvement.
    \item If no significant issues are identified, provide minimal or no suggestions for improvement.
  \end{itemize}
\end{itemize}

\textbf{Output Format:}
\begin{itemize}
  \item \textbf{experience\_summary}: A concise overview of the events at this node and the key outcomes.
  \item \textbf{confidence\_assessment}: High/Medium/Low with a recommendation for future exploration.
  \item \textbf{lessons\_learned}: Key takeaways or specific improvements based on the evaluation of the current node’s action.
  \item \textbf{comments}: Optional minor remarks, clarifications, or additional observations.
\end{itemize}

\end{promptbox}

\end{document}